\documentclass{article}

\usepackage{microtype}
\usepackage{graphicx}
\usepackage{subcaption}
\usepackage{booktabs} %
\usepackage{multirow}
\usepackage{colortbl} %
\usepackage{float}

\usepackage{hyperref}

\newif\ifarxiv
\arxivtrue

\ifarxiv
  \usepackage[preprint]{icml2026}
  \hypersetup{pdfsubject={}}
\else
  \usepackage{icml2026}
\fi
\usepackage{amsmath}
\usepackage{amssymb}
\usepackage{mathtools}
\usepackage{amsthm}

\usepackage[capitalize,noabbrev]{cleveref}

\theoremstyle{plain}

\theoremstyle{definition}

\theoremstyle{remark}

\usepackage[textsize=tiny]{todonotes}

\DeclareRobustCommand{\method}{%
  \mbox{P\textsuperscript{2}-MOPD}%
}
\icmltitlerunning{P\smash{\textsuperscript{2}}-MOPD: Accelerating Multi-Teacher On-Policy Distillation via Parameter Priors}

\newcommand{\Lt}{\mathcal{L}_t}
\newcommand{\sd}[1]{{\scriptsize$\pm$#1}} %
\newcommand{\supp}{\mathcal{V}}
\newcommand{\modset}{\mathcal{M}}
\newcommand{\sg}{\operatorname{sg}}
\newcommand{\Adam}{\operatorname{Adam}}
\newcommand{\AdamW}{\operatorname{AdamW}}

\begin{document}

\twocolumn[
  \icmltitle{\method{}: Accelerating Multi-Teacher On-Policy Distillation via Parameter Priors}

  \icmlsetsymbol{corr}{\textdagger}
  \icmlsetsymbol{lead}{\textdaggerdbl}

  \begin{icmlauthorlist}
    \icmlauthor{Sinuo Wang}{aiml,2}
    \icmlauthor{Jiacheng Lu}{2,4}
    \icmlauthor{Zeyu Zhang}{aiml}
    \icmlauthor{Maojia Song}{sutd}
    \icmlauthor{Rui Sun}{2,lead}
    \icmlauthor{Beidi Luan}{2}
    \icmlauthor{Yutong Xie}{mbzuai}
    \icmlauthor{Jing Li}{2}
    \icmlauthor{Daxin Jiang}{2}
    \icmlauthor{Qi Wu}{aiml,corr}
    \icmlauthor{Zuo Bai}{2,corr}
  \end{icmlauthorlist}

  \icmlaffiliation{aiml}{Australian Institute for Machine Learning, Adelaide University, Adelaide, Australia}
  \icmlaffiliation{2}{StepFun, Shanghai, China}
  \icmlaffiliation{sutd}{Singapore University of Technology and Design, Singapore }
  \icmlaffiliation{mbzuai}{Mohamed bin Zayed University of Artificial Intelligence, Abu Dhabi, United Arab Emirates}
  \icmlaffiliation{4}{School of Computer Science, Shanghai Jiao Tong University, Shanghai, China}

  \icmlcorrespondingauthor{Zuo Bai}{baizuo@stepfun.com}
  \icmlcorrespondingauthor{Qi Wu}{qi.wu01@adelaide.edu.au}

  \icmlkeywords{On-Policy Distillation, Model Merging, Task Arithmetic, Large Language Models}

  \vskip 0.3in
]

\printAffiliationsAndNotice{\textsuperscript{\textdagger}Co-corresponding authors. \textsuperscript{\textdaggerdbl}Project lead. }

\begin{abstract}
  Multi-teacher on-policy distillation consolidates expert capabilities by supervising student-generated trajectories, whereas task arithmetic offers a training-free alternative through linear parameter combinations. We view the linear combination of student and expert parameters as a single differentiable model and propose \method{}, which jointly optimizes the mixing coefficients and the student weights under the same on-policy distillation objective. Unlike task arithmetic, which stays within the span of the frozen student and expert parameters for any choice of merging coefficients, \method{} keeps the student trainable, so the experts act as \emph{parameter priors} without confining it to this span. Across four domains at 1.7B and 4B scales, \method{} achieves faster and more balanced recovery than MOPD from the same initialization, surpassing its final average recovery in half the training steps. At the end of training, our 1.7B model reaches 94.7\% average recovery, versus 80.9\% for MOPD.\ifarxiv{} Code is available at \url{https://github.com/SinuoWang/P2-MOPD}.\fi
\end{abstract}

\section{Introduction}

\begin{figure}[t]
\centering
\includegraphics[width=0.92\columnwidth]{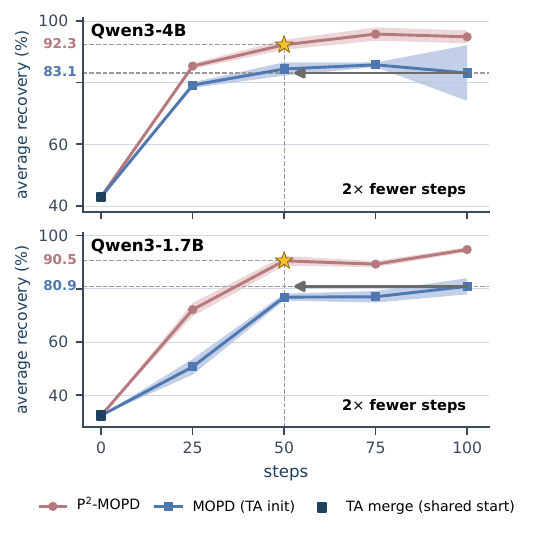}
\caption{Average recovery over training at Qwen3-4B (top) and Qwen3-1.7B (bottom). \method{} surpasses the final average recovery of MOPD (TA init) (arrow) after half the training steps (star). Mean over three seeds (shaded: $\pm1$ sd); both methods start from the same TA merge (square).}
\label{fig:main}
\end{figure}

Frontier large language models are increasingly built by training separate domain experts, often with domain-specific reinforcement learning~\cite{wangevolving}, and then consolidating them into a single generalist. Multi-teacher on-policy distillation (MOPD)~\cite{ma2026mopd}, in which each teacher supervises the student's own responses in its domain, has become a standard recipe for this step~\cite{xiaomi2026mimov2flash,xiaomi2026mimov26,deepseekai2026deepseekv4,yang2026nemotroncascade2,nvidia2026nemotron3ultra,kimiteam2026kimik3}. Ideally, the consolidated model retains the capability of every expert, rather than trading one domain for another. Because MOPD benefits from supervising the student's own behavior, its pipeline is heavy. Every step samples rollouts from the student and runs each teacher over them with teacher forcing to obtain token-level labels, and the student must approach each teacher's performance through many such steps. Prior work has also found that MOPD does not recover all domains evenly~\cite{kim2026no,gao2026openmopd}. Even when it starts from a merge of the experts, it can leave one domain far behind, and continued distillation can erode a capability that the merge already had. The other common way to consolidate experts is model merging. Task arithmetic~\cite{ilharco2023task} adds a weighted sum of the experts' task vectors to the base model without any training. Because the behavior of the merged model is never supervised, its quality is not guaranteed. In our experiments, for instance, a uniform merge recovers less than half of the gap between the base model and the teachers (\cref{tab:main}).

Task arithmetic falls short on its own, but its partial success is informative. If a training-free linear combination of the experts already recovers part of their capabilities, their parameters encode a useful prior for the consolidated model. We take this observation one step further. A linear combination of the base student and the frozen experts is itself a single model, differentiable in both its mixing coefficients and the student's weights, and can therefore be trained end to end under the MOPD objective. Viewed this way, the experts no longer serve only as an initialization or a one-off merge but act as parameter priors throughout training, so the student need not climb the whole way to each teacher on its own. Based on this view, we propose \method{} (Parameter-Prior MOPD; \cref{fig:overview}). It keeps the MOPD loss and its supervision on the student's own trajectories, but parameterizes the student as a linear combination of its own trainable weights and the frozen expert parameters (\cref{eq:param}). The student thus receives two kinds of guidance from the experts under a single loss: their logits supervise its outputs, as in MOPD, and their parameters enter its weights through the mixing coefficients, which the on-policy loss updates together with the student weights. Unlike task arithmetic, \method{} keeps the base student's weights fully trainable, so the consolidated model is not confined to the span of the student and expert parameters. Such confinement would be restrictive, since task vectors of separately trained experts interfere with each other~\cite{yadav2023ties,yu2024dare}, and with only $K$ directions per module the merged model can trade one domain against another but cannot make adjustments that none of the experts contains. In \method{}, the mixing coefficients move the model along expert-derived directions and the weights adapt it beyond them. The closest prior work, IM-MOPD~\cite{kim2026no}, also reuses the expert parameters but merges task vectors into the student at fixed points according to an external validation rule, whereas \method{} learns the combination continuously from the distillation gradient.

To understand why learning the combination helps, we project each model's change onto the task vectors (\cref{sec:mechanism}). MOPD from the merged start moves the student along most task vectors but hardly along that of the expert whose domain it leaves behind. \method{} moves an order of magnitude farther along this direction, and across methods the recovery of that domain tracks the movement. The learnable coefficients thus make reachable a teacher-aligned direction that student weight-only updates do not take, while the weights adapt the student outside the span of the task vectors, so the two channels divide the work.

Across four domains (medicine, law, finance and instruction following) at two scales, \method{} is both faster and more balanced than MOPD from the same merged start (\cref{sec:results}). After half the training steps, it already exceeds the final average recovery of MOPD at lower wall-clock cost (\cref{fig:main}), and at the end of training it raises worst-domain recovery from below half to nearly $90\%$. Learning the mixing coefficients alone, inside the span, falls far short, so both the expert prior and the freedom to leave its span are needed (\cref{sec:ablation}). Our contributions are as follows.
\begin{itemize}
\item We propose \method{}, which treats a linear combination of the student and frozen experts as a single model and trains its weights and mixing coefficients jointly under the MOPD objective (\cref{sec:method}).
\item Student weight-only MOPD updates barely move along the direction of the lagging expert; the learnable coefficients make this direction reachable, while the weights adapt the student beyond the task-vector span (\cref{sec:mechanism}).
\item At two scales, \method{} surpasses the final average recovery of MOPD in half the training steps and raises worst-domain recovery from below $50\%$ to nearly $90\%$ (\cref{sec:results,sec:ablation}).
\end{itemize}

\section{Related Work}
\label{sec:related}

\paragraph{Multi-Teacher On-Policy Distillation.}
On-policy distillation (OPD)~\cite{lu2025onpolicy,yang2025qwen3,yang2026gopd,li2026rethinking,fu2026rethinking,zhao2601self,xing2026trust,kang2026asyncopd} supervises the student with teacher feedback on its own trajectories and has become a dense-reward alternative to reinforcement learning in LLM post-training. MOPD~\cite{ma2026mopd} extends OPD to consolidating several RL-trained domain experts by routing each prompt to the teacher of its domain, and is used in recent frontier technical reports to consolidate domain experts into one model~\cite{xiaomi2026mimov2flash,xiaomi2026mimov26,deepseekai2026deepseekv4,yang2026nemotroncascade2,nvidia2026nemotron3ultra,kimiteam2026kimik3}. Follow-up work refines which teacher supervises which token and how strongly, through budget balancing across domains~\cite{gao2026openmopd}, label-free token routing~\cite{xu2026mopdrouter}, or soft-prompt teachers~\cite{ma2026promptsd}. These methods improve the supervision but treat the student as a plain set of weights. Closest to ours, IM-MOPD~\cite{kim2026no} shows that the expert parameters themselves also help: it initializes the student by task arithmetic and periodically merges task vectors of under-recovered domains back into it. However, these merges are rule-based and rely on an extra validation signal rather than the distillation gradient. \method{} instead treats the linear combination of trainable student parameters and frozen expert parameters as a single differentiable model, jointly optimizing the student weights and expert mixing coefficients throughout training under the same on-policy objective.

\paragraph{Model Merging.}
Model merging combines experts in weight space without training, for example by task arithmetic~\cite{ilharco2023task} and its interference-reducing variants~\cite{yadav2023ties,yu2024dare}. Because good fixed coefficients are hard to choose, another line learns them: AdaMerging~\cite{yang2024adamerging} fits task- or layer-wise coefficients by entropy minimization on unlabeled test data, and subsequent methods learn them from task data, expert logits or expert activations~\cite{zhang2024atlas,gauthiercaron2024dam,xu2025prodistill}. Other work interleaves merging with training as a discrete step between phases~\cite{choshen2022fusing,sun2025tinyr1,shen2025mot}. Together, these works show that trained expert parameters can be combined into a reasonable model even without training. They still treat merging as an operation that outputs a model, either offline or between training phases. Our view is different: since task arithmetic is a linear and fully differentiable function of its coefficients, the combination of student and expert parameters can be treated as one model. \method{} trains this model inside MOPD, so the expert parameters serve as a prior and the on-policy objective updates both the mixing coefficients and the student weights, without confining the student to the span of the task vectors.

\section{Method}
\label{sec:method}

\begin{figure*}[t]
\centering
\includegraphics[width=\textwidth,pagebox=mediabox,trim=15 12 2 55,clip]{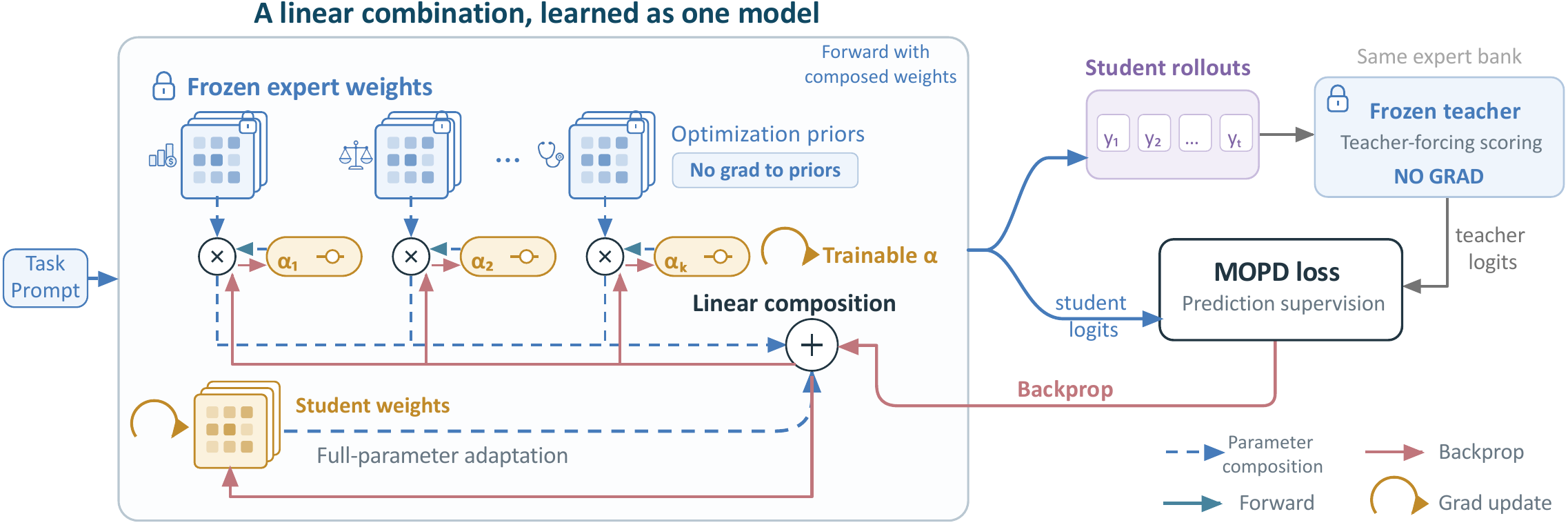}
\caption{Overview of \method{}. Each projection of the student is composed as its own learnable full weight plus a learnable linear combination of frozen expert directions, $W_m=B_m+\sum_k\alpha_{k,m}\tau_{k,m}$ (\cref{eq:param}), and the composed weights are used as one model for the forward pass. The student generates rollouts on prompts of all domains; the frozen teacher of each prompt's domain scores the student's tokens. The sampled tokens and the teacher logits are passed as detached data flow (gray lines), and the resulting MOPD loss is back-propagated to the student weights (student weight channel) and to the coefficients $\alpha$ (coefficient channel). No gradient reaches the experts or the teachers, which come from the same expert bank.}
\label{fig:overview}
\end{figure*}

\method{} rests on a simple view. A linear combination of model parameters is itself a single model, whose forward pass uses only the composed weights, and the composition is differentiable in its mixing coefficients. \method{} therefore reparameterizes the MOPD student as trainable student weights plus a learnable combination of frozen expert weights, and jointly optimizes the student weights and the mixing coefficients under the on-policy distillation objective (\cref{fig:overview}). The experts thus guide the student in two ways: their \emph{logits} supervise its behavior, and their \emph{weights} serve as parameter priors. \Cref{sec:method:setup} defines the notation and objective, \cref{sec:method:param} introduces the parameterization, and \cref{sec:method:update} presents the update rule and its geometric interpretation.

\subsection{Problem Formulation and Preliminaries}
\label{sec:method:setup}

\paragraph{Experts and task arithmetic.}
Let $\theta_0$ be a shared base model and $\theta_1,\dots,\theta_K$ be $K$ domain experts obtained from $\theta_0$ by domain-specific post-training. The \emph{task vector} of expert $k$ is $\tau_k=\theta_k-\theta_0$~\cite{ilharco2023task}, and a task-arithmetic (TA) merge is
\begin{equation}
\theta_{\mathrm{TA}}(\lambda)=\theta_0+\sum_{k=1}^{K}\lambda_k\,\tau_k ,
\label{eq:ta}
\end{equation}
with coefficients $\lambda\in\mathbb{R}^K$ (uniform, $\lambda_k=1/K$, throughout this paper). TA is training-free, but the merge is confined to the $K$-dimensional affine span of the task vectors.

\paragraph{Multi-teacher on-policy distillation.}
MOPD trains one student $\pi_\theta$ on prompts $x$ with known domain labels $d(x)\in\{1,\dots,K\}$. At step $t$ the current student $\pi_{\theta_t}$ samples a response $y$ for each prompt, and the expert of the prompt's domain, $\pi_{\theta_{d(x)}}$, scores it. Let $p(v)$ and $q(v)$ denote the student's and the routed teacher's next-token distributions at position $i$. Following the top-$k$ form of MOPD~\cite{ma2026mopd} as implemented in Open-MOPD~\cite{gao2026openmopd}, each candidate $v$ in a student-defined support $\supp_i$ (top-$S$ student tokens with a nucleus cut; \cref{app:objective}) receives the reward
\begin{equation}
r_i(v)=-\,w_i(v)\,\big[\log p(v)-\log q(v)\big],\qquad v\in\supp_i ,
\label{eq:reward}
\end{equation}
with $w_i(v)=p(v)/\sum_{u\in\supp_i}p(u)$, a support-restricted reverse-KL signal (\cref{app:objective}). The rewards are treated as constants ($\sg$), and the student is updated with a clipped policy-gradient surrogate. Since every rollout batch is used for one optimizer update, the update direction is the gradient of
\begin{equation}
\begin{split}
\Lt(\theta)=-\frac{1}{N_t}&\sum_{(x,y)\in\mathcal{B}_t}\ \sum_{i\in y}\ \sum_{v\in\supp_i}\\
&\sg\!\big[r_i(v)\big]\,\log\pi_\theta(v\mid x,y_{<i})
\end{split}
\label{eq:opd}
\end{equation}
at $\theta=\theta_t$, where $\mathcal{B}_t$ is the rollout batch and $N_t$ its number of unmasked response tokens. Standard MOPD applies this gradient to all parameters of a student initialized at $\theta_0$, at an SFT warm-up, or at the TA merge $\theta_{\mathrm{TA}}(\lambda)$, which we denote MOPD, MOPD (SFT warm-up) and MOPD (TA init)~\cite{ma2026mopd,kim2026no}; in MOPD (TA init) the task vectors enter only through the initial point.

\subsection{Expert Parameters as Differentiable Priors}
\label{sec:method:param}

\paragraph{Parameterization.}
\method{} is applied module-wise. A module $m\in\modset$ is one linear projection of a decoder layer (query, key, value and output in attention; gate, up and down in the MLP). The remaining tensors $\phi$ (token embeddings, normalization weights and the output head, including a tied embedding) are trained as in MOPD. For a module $m\in\modset$ let $W^{(k)}_m$ be the corresponding tensor of expert $k$, with task vector $\tau_{k,m}=W^{(k)}_m-W^{(0)}_m$. \method{} parameterizes the student's module weights as
\begin{equation}
W_m(B,\alpha)=B_m+\sum_{k=1}^{K}\alpha_{k,m}\,\tau_{k,m},\qquad m\in\modset,
\label{eq:param}
\end{equation}
where the full tensors $B=\{B_m\}_{m\in\modset}$ and the scalar mixing coefficients $\alpha\in\mathbb{R}^{K\times|\modset|}$ are both trainable and the task vectors are frozen (left block of \cref{fig:overview}). No gradient is back-propagated into any teacher network: the teachers enter the loss only through the constant rewards, and the task vectors enter the parameterization only as fixed directions. Since $B$ is unconstrained, \cref{eq:param} represents exactly the same models as a plain student. What it changes is how the student moves: a single scalar step on $\alpha_{k,m}$ moves the whole module along a task vector. In this sense the frozen experts act as \emph{parameter priors}, a bias on the optimization path rather than on the model class (\cref{sec:method:update}).

\paragraph{Coefficient granularity.}
\Cref{eq:param} gives every module its own $K$ coefficients (\emph{module} granularity, $K|\modset|$ scalars). The coefficients can also be tied across modules: at \emph{layer} granularity all projections of one decoder layer share $K$ coefficients, and at \emph{model} granularity there is one coefficient per expert ($K$ scalars), the learnable analogue of \cref{eq:ta}. Tying only constrains $\alpha$; every module keeps its own task vectors and weights. It is implemented by writing the full table as $\alpha=\tilde\alpha\,T^{\top}$ for a fixed binary assignment matrix $T$ and optimizing the free parameters $\tilde\alpha$, so the gradient of a tied coefficient is the \emph{sum} of the gradients of the modules it covers. Our main experiments use module granularity; the method does not depend on it, and finer tying is not necessarily better (\cref{sec:ablation,tab:granularity}).

\paragraph{Initialization.}
\method{} starts from the TA merge $\theta_{\mathrm{TA}}(\lambda)$, in which \emph{every} tensor (modules and $\phi$ alike) is merged with the TA coefficients $\lambda$. It sets $B_m=W^{(0)}_m$ and $\alpha_{k,m}=\lambda_k$, so that $W_m(B,\alpha)$ reproduces the merged module weights exactly, and takes $\phi$ from the merge. Equivalently, at initialization $W_m=(1-\sum_k\alpha_{k,m})\,W^{(0)}_m+\sum_k\alpha_{k,m}\,W^{(k)}_m$: a linear combination of the base and expert weights whose coefficients remain learnable for the entire run.

\paragraph{Gradients.}
Let $G_m=\partial\Lt/\partial W_m$ be the ordinary gradient of the distillation loss with respect to the composed weight, evaluated at the current weights. By the chain rule through \cref{eq:param},
\begin{equation}
\frac{\partial\Lt}{\partial B_m}=G_m,
\qquad
\frac{\partial\Lt}{\partial\alpha_{k,m}}=\big\langle G_m,\,\tau_{k,m}\big\rangle .
\label{eq:grad}
\end{equation}
The coefficient gradient is the directional derivative of the loss along the frozen task vector: a negative value means that, to first order, increasing module $m$'s contribution from expert $k$ decreases the current batch loss. Because the batch contains prompts of all domains, every coefficient receives a sum of contributions from all domains, $\langle G_m,\tau_{k,m}\rangle=\sum_d\langle G^{(d)}_m,\tau_{k,m}\rangle$, not only from the domain of expert $k$.

\subsection{Joint Optimization and Its Interpretation}
\label{sec:method:update}

\begin{algorithm}[tb]
\caption{\method{}: one training step (module granularity)}
\label{alg:p2mopd}
\begin{algorithmic}[1]
\REQUIRE merged start $\theta_{\mathrm{TA}}(\lambda)$; frozen task vectors $\tau_{k,m}$; learning rates $\eta_W,\eta_\alpha$; clip norms $c_W,c_\alpha$
\STATE $W\leftarrow$ module tensors of $\theta_{\mathrm{TA}}(\lambda)$;\ $\phi\leftarrow$ its remaining tensors;\ $\alpha_{k,m}\leftarrow\lambda_k$
\FOR{$t=0,1,\dots$}
  \STATE Sample prompts $x$ with labels $d(x)$; generate $y\sim\pi_{\theta_t}(\cdot\mid x)$
  \STATE Score $(x,y)$ with the routed teacher $\pi_{\theta_{d(x)}}$; form $r_i(v)$ \hfill\cref{eq:reward}
  \STATE $(G,G_\phi)\leftarrow\nabla_{(W,\phi)}\Lt$ at $\theta_t$ \hfill\cref{eq:opd}
  \STATE $g_{k,m}\leftarrow\langle G_m,\tau_{k,m}\rangle$ from the \emph{unclipped} $G$ \hfill\cref{eq:grad}
  \STATE Clip $(G,G_\phi)$ to global norm $c_W$; $(W^{+},\phi_{t+1})\leftarrow\AdamW_{\eta_W}\big((W,\phi),(G,G_\phi)\big)$
  \STATE Clip $g$ to norm $c_\alpha$; $\alpha_{t+1}\leftarrow\Adam_{\eta_\alpha}(\alpha_t,g)$
  \STATE $W_{t+1,m}\leftarrow W^{+}_m+\sum_{k}(\alpha_{t+1}-\alpha_t)_{k,m}\,\tau_{k,m}$ \hfill\cref{eq:update}
  \STATE Synchronize $(W_{t+1},\phi_{t+1})$ to the rollout engine
\ENDFOR
\ENSURE a standard checkpoint $(W,\phi)$ with $\alpha$ folded in; the trajectory of $\alpha$ for analysis
\end{algorithmic}
\end{algorithm}

\paragraph{Update rule.}
\Cref{eq:param} is the conceptual model: it defines $\alpha$ and gives the coefficient gradient through the chain rule. The optimizer, however, does not maintain $B$. It maintains the materialized weights $W$ in full precision and advances them with a two-channel step (\cref{alg:p2mopd}; the two backward paths of \cref{fig:overview}):
\begin{equation}
\begin{split}
W_{t+1}={}&\AdamW_{\eta_W}\!\big(W_t,\,\mathrm{clip}_{c_W}(G_t)\big)\\
&+\sum_{k}\big(\alpha_{t+1}-\alpha_t\big)_{k}\,\tau_{k},
\end{split}
\label{eq:update}
\end{equation}
where $\alpha_{t+1}=\Adam_{\eta_\alpha}\big(\alpha_t,\mathrm{clip}_{c_\alpha}(D^{\top}G_t)\big)$ and $D^{\top}G_t$ collects the inner products of \cref{eq:grad}. Three details of \cref{eq:update} matter. (i)~Both channels consume the \emph{same} pre-update gradient $G_t$, computed once at $\theta_t$; there is no second backward pass. (ii)~The coefficient gradient is taken from $G_t$ \emph{before} the weight gradient is clipped, and the two channels are clipped separately, so neither channel's clipping budget is consumed by the other. (iii)~The student weight channel uses no weight decay, so running AdamW on the materialized $W$ and adding the coefficient increment is identical to running the same optimizer on $B$ and recomposing $W=B+\sum_k\alpha_k\tau_k$, because $G_t$ is the same for $W$ and $B$ by \cref{eq:grad} (\cref{app:update}). The non-$\modset$ parameters $\phi$ follow the AdamW channel alone. The per-step displacement of a module is therefore
\begin{equation}
\Delta W_m=\underbrace{\Delta^{\AdamW}_m}_{\text{student weight channel}}+\underbrace{\textstyle\sum_{k}\Delta\alpha_{k,m}\,\tau_{k,m}}_{\in\,\mathrm{span}\{\tau_{1,m},\dots,\tau_{K,m}\}} .
\label{eq:displacement}
\end{equation}
At module granularity the second term lies, for each module, in a subspace of dimension at most $K$; over the whole network the coefficient channel moves within a structured subspace of dimension at most $K|\modset|$, while the first term is unrestricted and moves both inside and outside that subspace. Coarser tying (layer, model) replaces the per-module span by the smaller span of stacked task vectors, dimension at most $KL$ or $K$.

\paragraph{Parameter priors as preconditioning.}
Like reparameterizations that leave the model class unchanged but alter the optimization geometry~\cite{salimans2016weight,arora2018optimization}, the two channels can be read as one preconditioned step~\cite{amari1998natural}. Fix one module, write $w=b+D\alpha$ with $D=[\mathrm{vec}\,\tau_{1},\dots,\mathrm{vec}\,\tau_{K}]$, and let $g=\nabla_w\Lt$ for a fixed batch. If both channels used plain SGD, with no momentum, no clipping and no decay, then $\Delta b=-\eta_W g$ and $\Delta\alpha=-\eta_\alpha D^{\top}g$, hence
\begin{equation}
w^{+}=w-\big(\eta_W I+\eta_\alpha DD^{\top}\big)\,g .
\label{eq:precond}
\end{equation}
The parameter prior thus acts as a preconditioner: a positive semidefinite term of rank at most $K$ amplifies the gradient component along the task vectors, with strength set by $\eta_\alpha$ relative to $\eta_W$ (under layer or model tying, $D$ is replaced by the stacked task vectors of the tied group). Adam and clipping act on each channel separately, so \cref{eq:precond} describes the direction of the change, not its size; \cref{sec:experiments} measures its effect.

\section{Experiments}
\label{sec:experiments}

\subsection{Experimental Setup}
\label{sec:setup}

\paragraph{Models and data.}
We follow the protocol of IM-MOPD~\cite{kim2026no} at two scales, Qwen3-4B and Qwen3-1.7B. The base $\theta_0$ is the general SFT reference model of that work, and at each scale four domain teachers (medical, law, finance, instruction following) are fine-tuned from the same $\theta_0$, so the task vectors $\tau_k=\theta_k-\theta_0$ are well defined. The domains are evaluated on MedQA, CaseHOLD, FinQA and IFBench. We omit the fifth domain, $\tau^2$-bench~\cite{barres2025tau2}, because its user simulator requires a paid commercial LLM API. Every training batch contains 128 prompts, 32 from each domain, and each prompt is distilled from the teacher of its domain. Teacher provenance, prompt pools and test splits are given in \cref{app:setup}.

\paragraph{Methods.}
\emph{MOPD (TA init)} is the primary control: MOPD from the uniform TA merge ($\lambda_k=0.25$), that is, \method{} with the mixing coefficients $\alpha$ frozen at $\lambda$ (the $\eta_\alpha=0$ corner of \cref{alg:p2mopd}). \emph{MOPD} starts from $\theta_0$ and \emph{MOPD (SFT warm-up)} from an SFT warm-up on teacher samples. \emph{Open-MOPD}~\cite{gao2026openmopd} adds per-domain budget balancing, and \emph{IM-MOPD}~\cite{kim2026no} periodically merges the task vectors of under-recovered domains into the student; both start from the same merge. \method{} uses module-wise granularity unless stated otherwise.

\paragraph{Training and evaluation.}
All experiments run on eight GPUs (\cref{app:hparams}). Every run trains for 100 steps of one rollout batch each, the training budget of IM-MOPD; we call the checkpoint at step 100 the \emph{final} checkpoint. Checkpoints at steps 25, 50, 75 and 100 are scored with the official IM-MOPD evaluation harness, which samples three responses per test item with evaluation seeds 42, 43 and 44 and averages their scores (avg@3; \cref{app:eval}). Per-domain \emph{recovery} is $(s-s_{\mathrm{base}})/(s_{\mathrm{teacher}}-s_{\mathrm{base}})$; the headline metric is the \emph{average recovery} over the four domains, and the \emph{worst-domain recovery} measures balance. Because the base-to-teacher spans differ by up to a factor of three across domains, we report raw scores alongside. \method{}, MOPD (TA init) and IM-MOPD are also trained with three training seeds at both scales; we report their means and standard deviations over the three training runs and compare \method{} and MOPD (TA init) in pairs of runs with the same training seed.

\subsection{Main Results}
\label{sec:results}

\begin{table*}[t]
\centering
\caption{Final results. Per-domain test scores (\%, avg@3: each item is scored by the mean over three samples at temperature $1.0$) and summary metrics; for \method{}, MOPD (TA init) and IM-MOPD all entries are three-seed means and the summary metrics show $\pm$ sd. Avg. Recovery: average recovery over the four domains; Avg. Score: mean raw score over the four benchmarks; Worst: worst-domain recovery; AULC: area under the learning curve, i.e.\ the average recovery averaged over the checkpoints at 25, 50, 75 and 100 steps. The base student and the teachers define recovery $0$ and $100$; the TA merge is the untrained common start. Bold: best trained method.}
\label{tab:main}
\footnotesize
\setlength{\tabcolsep}{4pt}
\begin{tabular}{l cccc | cccc}
\toprule
Method & MedQA & CaseHOLD & FinQA & IFBench & Avg. Recovery & Avg. Score & Worst & AULC \\
\midrule
\rowcolor[RGB]{225, 235, 245}\multicolumn{9}{c}{Qwen3-4B} \\
Base student ($\theta_0$) & 68.84 & 62.03 & 58.76 & 25.89 & 0 & 53.88 & -- & -- \\
Domain teachers & 79.86 & 73.57 & 75.41 & 59.22 & 100 & 72.02 & -- & -- \\
\midrule
TA merge & 73.95 & 66.00 & 70.01 & 33.78 & 42.98 & 60.93 & 23.67 & -- \\
MOPD & 62.90 & 66.30 & 75.04 & \textbf{61.33} & 46.80 & 66.39 & $-53.94$ & 20.43 \\
MOPD (SFT warm-up) & 69.76 & 71.47 & 74.22 & 53.11 & 66.15 & 67.14 & 8.32 & 61.86 \\
MOPD (TA init) & 76.07 & \textbf{72.62} & \textbf{75.35} & 51.08 & 83.13\sd{9.03} & 68.78\sd{3.67} & 47.22\sd{23.72} & 83.13\sd{2.72} \\
Open-MOPD & 75.07 & 72.26 & 73.73 & 59.78 & 84.20 & 70.21 & 56.55 & 80.97 \\
IM-MOPD & 78.22 & 71.76 & 73.25 & 57.78 & 88.02\sd{1.33} & 70.25\sd{0.32} & 84.12\sd{0.20} & 79.11\sd{2.83} \\
\rowcolor{gray!15}\textbf{\method{}} & \textbf{78.74} & 72.58 & 75.01 & 59.41 & \textbf{94.85}\sd{2.06} & \textbf{71.43}\sd{0.46} & \textbf{89.42}\sd{2.54} & \textbf{92.05}\sd{1.16} \\
\midrule
\rowcolor[RGB]{225, 235, 245}\multicolumn{9}{c}{Qwen3-1.7B} \\
Base student ($\theta_0$) & 44.10 & 44.65 & 43.62 & 19.89 & 0 & 38.07 & -- & -- \\
Domain teachers & 58.99 & 70.52 & 62.66 & 46.44 & 100 & 59.65 & -- & -- \\
\midrule
TA merge & 50.67 & 57.20 & 47.40 & 24.44 & 32.41 & 44.93 & 17.15 & -- \\
MOPD & 5.21 & 68.10 & 53.59 & 34.56 & $-15.74$ & 40.36 & $-261.18$ & $-1.55$ \\
MOPD (SFT warm-up) & 38.33 & 68.02 & 51.87 & 34.11 & 37.13 & 48.08 & $-38.72$ & 30.04 \\
MOPD (TA init) & 48.87 & 69.15 & \textbf{62.96} & 45.22 & 80.93\sd{3.02} & 56.55\sd{0.52} & 32.03\sd{10.12} & 71.38\sd{2.04} \\
Open-MOPD & 50.80 & 69.08 & 60.80 & 43.67 & 79.79 & 56.08 & 44.99 & 73.26 \\
IM-MOPD & 55.22 & 62.49 & 54.71 & 38.30 & 67.81\sd{9.18} & 52.68\sd{2.10} & 48.73\sd{17.02} & 53.17\sd{1.55} \\
\rowcolor{gray!15}\textbf{\method{}} & \textbf{57.20} & \textbf{69.49} & 61.53 & \textbf{46.66} & \textbf{94.74}\sd{0.91} & \textbf{58.72}\sd{0.21} & \textbf{88.01}\sd{1.27} & \textbf{86.68}\sd{0.63} \\
\bottomrule
\end{tabular}
\end{table*}

\paragraph{Faster recovery.}
\Cref{fig:main} compares \method{} with MOPD (TA init), which differs from it only in freezing the mixing coefficients $\alpha$ at their TA values. \method{} is ahead from the first checkpoint, by $+6.2$ (4B) and $+21.5$ (1.7B) average-recovery points at step 25, and stays ahead in all 24 seed-paired comparisons (two scales, three seeds, four checkpoints; \cref{app:eval}). After half the training steps it reaches $92.3$ (4B) and $90.5$ (1.7B), already above the final average recovery of MOPD (TA init) ($83.1$ and $80.9$) in every seed pair, and the advantage carries over to wall-clock time (\cref{sec:cost}). The two scales differ in how the gap evolves. At 4B it widens at every checkpoint, to $+11.7$ at the end of training. At 1.7B it emerges at the first checkpoint, and \method{} continues to recover gradually.

\paragraph{Final performance.}
At the end of training \method{} leads at both scales in average recovery (\cref{tab:main}) and in mean raw score, by $+2.66$ (4B) and $+2.17$ (1.7B) points (paired, three seeds); the raw gain is positive in every seed. \method{} also ends well above MOPD and MOPD (SFT warm-up), and above IM-MOPD at both scales, with non-overlapping seed ranges ($93.5$--$97.2$ versus $86.6$--$89.2$ at 4B, $94.2$--$95.8$ versus $59.0$--$77.4$ at 1.7B). At 1.7B, IM-MOPD's end point depends on which domains its validation-gated rule happens to merge, and its average recovery varies by $18$ points across seeds.

\paragraph{Balanced recovery.}
\begin{figure}[!t]
\centering
\includegraphics[width=\columnwidth]{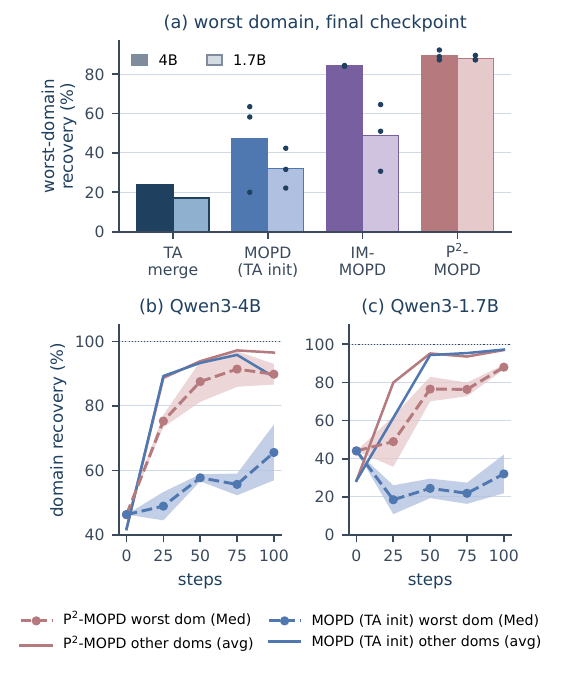}
\caption{Balance of recovery across domains. (a) Final worst-domain recovery (bars: three-seed mean; dots: seeds). (b,\,c) Recovery of the worst domain, MedQA (dashed; mean $\pm1$ sd over three seeds), and the mean recovery of the other three domains (solid). MedQA is the lowest domain for both methods at almost every checkpoint. MOPD (TA init) recovers the other three domains as fast as \method{} but leaves MedQA behind.}
\label{fig:balance}
\end{figure}
The clearest effect is on the worst-recovered domain (\cref{fig:balance}). For MOPD (TA init) this is almost always MedQA, while the other three domains are recovered about as fast as by \method{} (\cref{fig:balance}b,\,c). At 1.7B, MOPD (TA init) even ends below the MedQA recovery of the TA start ($44\%$) in all three seeds: continued distillation erodes a capability the merge already had. \method{} closes this gap from the first checkpoint and ends with a worst-domain recovery of $89.4$ (4B) and $88.0$ (1.7B), against $47.2$ and $32.0$ for MOPD (TA init), while the across-domain std of recovery falls from $21.8$ to $4.6$ (4B) and from $28.4$ to $4.8$ (1.7B). IM-MOPD is also balanced at 4B (worst domain $84.1$), where every seed merges the medical teacher at step 25, but it first exceeds $80\%$ average recovery only at step 50 or 75, against step 25 for \method{}; at 1.7B its worst domain is $48.7$.

\paragraph{Where the gain comes from.}
\begin{table}[t]
\centering
\caption{Per-domain recovery (\%) of \method{} and its difference from MOPD (TA init). Final: recovery of \method{} at the final checkpoint (three-seed mean). $\Delta$: mean difference, \method{} minus MOPD (TA init), with sd over all twelve seed-paired comparisons per scale (four checkpoints $\times$ three seeds).}
\label{tab:domain_diff}
\small
\setlength{\tabcolsep}{5pt}
\begin{tabular}{l cc cc}
\toprule
& \multicolumn{2}{c}{Qwen3-4B} & \multicolumn{2}{c}{Qwen3-1.7B} \\
\cmidrule(lr){2-3}\cmidrule(lr){4-5}
Domain & Final & $\Delta$ & Final & $\Delta$ \\
\midrule
MedQA & 89.8 & $+29.0$\sd{8.3} & 88.0 & $+48.2$\sd{14.4} \\
CaseHOLD & 91.4 & $+3.8$\sd{4.2} & 96.0 & $+4.5$\sd{3.1} \\
FinQA & 97.6 & $-1.9$\sd{3.5} & 94.1 & $+0.8$\sd{15.6} \\
IFBench & 100.6 & $+4.8$\sd{23.2} & 100.9 & $+7.6$\sd{12.9} \\
\midrule
Avg. Recovery & 94.9 & $+8.9$\sd{4.6} & 94.7 & $+15.3$\sd{4.5} \\
\bottomrule
\end{tabular}
\end{table}

\Cref{tab:domain_diff} breaks the average-recovery difference down by domain. The gain is concentrated where MOPD (TA init) falls short: MedQA is higher in every seed-paired comparison at both scales, by $+29.0$ (4B) and $+48.2$ (1.7B) recovery points on average, several times the difference in any other domain. CaseHOLD is mostly higher by small margins. On FinQA, which MOPD (TA init) already recovers almost fully, \method{} is slightly lower in most pairs, and IFBench differs in both directions with a large spread across pairs. These differences lie around a high level: at the end of training \method{} recovers on average at least $88\%$ of the teacher gap in every domain, including $98\%$ (4B) and $94\%$ (1.7B) on FinQA (\cref{tab:main}), and the small FinQA deficit reflects the competition for $\alpha_{\mathrm{med}}$ analysed in \cref{sec:mechanism}. \method{} thus lifts the domain that MOPD leaves behind while keeping every other domain close to its teacher.

\subsection{Why Learnable Coefficients Help}
\label{sec:mechanism}

\begin{figure}[t]
\centering
\includegraphics[width=\columnwidth]{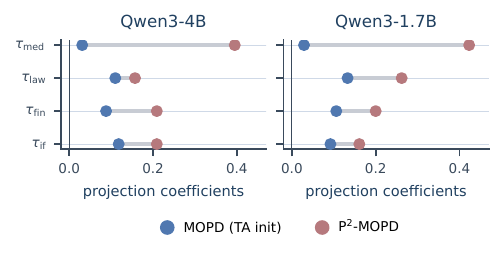}
\caption{Task-vector projection of the final displacement from the TA start in $\alpha$ units (three-seed means; seed sd $\le0.007$). Each line connects MOPD (TA init) to \method{} along one task vector. The share of the displacement energy that lies in the task-vector span is $5.2\%$ / $10.7\%$ for MOPD (TA init) and $85.4\%$ / $90.0\%$ for \method{} (4B / 1.7B).}
\label{fig:geometry}
\end{figure}

\paragraph{Task-vector projection.}
To see where each method moves the student, we project the displacement of every module from the TA start onto the span of its four task vectors, in units of $\alpha$ (\cref{app:setup}). MOPD (TA init) does use the task vectors, but unevenly: it moves $+0.09$ to $+0.13$ along $\tau_{\mathrm{law}}$, $\tau_{\mathrm{fin}}$ and $\tau_{\mathrm{if}}$, and only $+0.03$ along $\tau_{\mathrm{med}}$, the direction of the domain it fails to recover, at both scales and in every seed (\cref{fig:geometry}). \method{} moves farthest along exactly this direction, by $+0.40$ (4B) and $+0.42$ (1.7B), 13--15 times as far. Within \method{}, the two channels divide the work: the coefficient channel accounts for $81\%$ (4B) and $84\%$ (1.7B) of the displacement energy, whereas the student weight channel moves almost entirely outside the span (only $1.7\%$ and $4.0\%$ of its displacement lies inside). The same projection also accounts for MedQA across methods. Among the final checkpoints of \method{}, MOPD (TA init) and IM-MOPD at both scales, the three that recover MedQA above $84\%$ all moved at least $0.30$ along $\tau_{\mathrm{med}}$: \method{} at both scales through its coefficients, and 4B IM-MOPD through merging the medical teacher at step 25. The other three moved at most $0.03$: MOPD (TA init) at both scales, and 1.7B IM-MOPD, which did not merge the medical teacher. Recovering the lagging domain thus coincides with a large step along its task vector, a step that student weight-only updates do not take and that \method{} takes through one coefficient per module.

\paragraph{Conflict resolution in coefficient space.}
\method{} turns cross-domain interference into a contest over a few interpretable scalars, settled continuously by the gradient in favour of the lagging domain. The per-domain coefficient gradients show where the conflict lives: the medical gradient lies 98\% on $\alpha_{\mathrm{med}}$ and has cosine $-0.78$ / $-0.63$ with the finance / IF gradients at 4B ($-0.85$ / $-0.81$ at 1.7B), so the domains disagree mainly about one scalar per module. They also show how it is settled: at every step the medical gradient pushes $\alpha_{\mathrm{med}}$ up while the finance and IF gradients push it down, their sum follows the medical gradient, and $\alpha_{\mathrm{med}}$ rises from $0.25$ to $0.63$ (4B) and $0.65$ (1.7B). The price is paid by the domains that lose this contest, and it is small: a slight FinQA deficit and mixed IFBench differences (\cref{tab:domain_diff}), against a far more balanced recovery overall (\cref{fig:balance}). \emph{The student weight channel is what makes this resolution possible.} With the weights frozen, the coefficients alone must serve all four domains and do not favour $\alpha_{\mathrm{med}}$ (its lead over the mean of the other three is $+0.06$ / $-0.07$, against $+0.20$ / $+0.22$ in \method{}); with the weights free to adapt the other domains, the coefficients can specialize on $\tau_{\mathrm{med}}$. The conflict is real but does not explain the gap of MOPD (TA init), which we test by training on medical prompts alone (\cref{tab:mechanism}, 1.7B). Without the other domains, MOPD (TA init) still barely moves along $\tau_{\mathrm{med}}$ ($+0.048$, against $+0.028$ with four domains), its medical KL stays more than seven times that of \method{} under the same change, and its MedQA recovery falls to $-6.9$. Inside \method{}, removing the other domains adds only 3--9 points of MedQA recovery from step 50 on, small against the gap of 65 points to MOPD (TA init).

\begin{table}[t]
\centering
\caption{Controlled interventions (1.7B, final checkpoint). Each indented row changes one factor of the run above it. $\alpha_{\mathrm{med}}$: whether a learnable coefficient moves the student along $\tau_{\mathrm{med}}$. Med KL: training reverse KL to the medical teacher on student rollouts (mean of steps 91--100). MedQA and Avg.: MedQA recovery and average recovery (\%). --: not reported.}
\label{tab:mechanism}
\small
\setlength{\tabcolsep}{3.5pt}
\begin{tabular}{lcccc}
\toprule
Run & $\alpha_{\mathrm{med}}$ & Med KL$\downarrow$ & MedQA$\uparrow$ & Avg.$\uparrow$ \\
\midrule
MOPD (TA init) & $\times$ & 0.115 & 22.1 & 78.1 \\
\quad medical prompts only & $\times$ & 0.098 & $-6.9$ & -- \\
\midrule
\method{} & \checkmark & 0.032 & 87.2 & \textbf{95.8} \\
\quad medical prompts only & \checkmark & \textbf{0.013} & 90.2 & -- \\
\quad random directions & $\times$ & 0.111 & 1.4 & 70.2 \\
\quad learn only $\alpha_{\mathrm{med}}$ & \checkmark & 0.028 & \textbf{94.1} & 95.0 \\
\quad freeze $\alpha_{\mathrm{med}}$ & $\times$ & 0.112 & 24.2 & -- \\
\bottomrule
\end{tabular}
\end{table}

\paragraph{Isolating the medical direction.}
The remaining explanation is reachability: MOPD (TA init) falls short on MedQA because student weight-only updates do not move along $\tau_{\mathrm{med}}$. The other interventions in \cref{tab:mechanism} separate this direction from the extra capacity that the coefficients add. If the gain came from extra trainable directions, any $K$ directions per module would serve; random directions in place of the task vectors do not beat MOPD (TA init) at any checkpoint. Learning only the $\alpha_{\mathrm{med}}$ row, one scalar per module, recovers nearly all of the gain ($95.0$ versus $95.8$ average recovery), and freezing $\alpha_{\mathrm{med}}$ returns MedQA to the level of MOPD (TA init) ($24.2$ versus $22.1$), so the gain follows the medical coefficient in both directions. Learning only $\alpha_{\mathrm{med}}$ requires knowing in advance which domain lags; \method{} learns all coefficients and finds this direction on its own. The learnable coefficients thus make reachable a teacher-aligned direction that student weight-only updates do not take, while the student weight channel adapts the other domains.

\subsection{Training Cost}
\label{sec:cost}

\begin{table}[t]
\centering
\caption{Cost to reach a fixed quality on $8$ GPUs. Step: mean seconds per step. Cost: wall-clock of 100 steps relative to MOPD (TA init). Time to target: wall-clock to exceed the final average recovery of MOPD (TA init), as a fraction of its 100-step wall-clock; checkpoints are 25 steps apart, so these are upper bounds. Recovery at budget: average recovery of the last checkpoint within the wall-clock of 100 steps of MOPD (TA init), which is step 75 for \method{}. Recoveries are three-seed means; all granularities are in \cref{tab:cost_full}.}
\label{tab:cost}
\small
\setlength{\tabcolsep}{3pt}
\begin{tabular}{llcccc}
\toprule
& Method & Step (s) & Cost & \begin{tabular}[b]{@{}c@{}}Time to\\target\end{tabular} & \begin{tabular}[b]{@{}c@{}}Recovery\\at budget\end{tabular} \\
\midrule
\multirow{2}{*}{4B} & MOPD (TA init) & 180.6 & $1.00\times$ & $1.00\times$ & 83.1 \\
& \cellcolor{gray!15}\textbf{\method{}} & \cellcolor{gray!15}225.6 & \cellcolor{gray!15}$1.24\times$ & \cellcolor{gray!15}$\mathbf{\le 0.31\times}$ & \cellcolor{gray!15}\textbf{95.7} \\
\midrule
\multirow{2}{*}{1.7B} & MOPD (TA init) & 153.9 & $1.00\times$ & $1.00\times$ & 80.9 \\
& \cellcolor{gray!15}\textbf{\method{}} & \cellcolor{gray!15}167.4 & \cellcolor{gray!15}$1.09\times$ & \cellcolor{gray!15}$\mathbf{\le 0.55\times}$ & \cellcolor{gray!15}\textbf{89.3} \\
\bottomrule
\end{tabular}
\end{table}

\method{} adds only the coefficient table and its optimizer state (at most on the order of $10^3$ scalars), and its output is a standard checkpoint with $\alpha$ folded into the weights, so inference is unchanged. Training makes each step more expensive but needs far fewer steps. The extra time comes from streaming the task vectors to compute the coefficient gradients and the composed update (\cref{app:update}). This lengthens the actor update from $14.7$ to $73.8$\,s at 4B and from $11.1$ to $37.1$\,s at 1.7B. However, it is rollout generation, not the actor update, that dominates each step, so 100 steps of \method{} take only $1.24\times$ (4B) and $1.09\times$ (1.7B) the wall-clock of MOPD (TA init). \Cref{tab:cost} converts this into the time needed to reach a fixed quality. MOPD (TA init) reaches its final average recovery ($83.1$ / $80.9$) only at the end of its 100 steps. \method{} already exceeds it at step 25 at 4B ($85.4$, $0.31\times$ the wall-clock) and at step 50 at 1.7B ($90.5$, $0.55\times$); since checkpoints are 25 steps apart, these are upper bounds. Within the wall-clock of 100 steps of MOPD (TA init), \method{} completes about 80 (4B) and 92 (1.7B) steps, and its step-75 checkpoint, which fits in this budget at both scales, reaches $95.7$ / $89.3$. From step 50 on, it also exceeds the best checkpoint of MOPD (TA init) at any step ($85.8$ / $80.9$).
\section{Ablation Studies}
\label{sec:ablation}

\begin{figure}[t]
\centering
\includegraphics[width=\columnwidth]{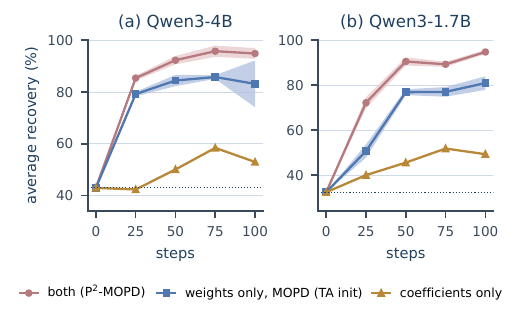}
\caption{Inside versus beyond the task-vector span (module granularity): mixing coefficients only (student weights frozen, confined to the span), student weights only (mixing coefficients frozen; MOPD (TA init)) and both (\method{}). The last two are three-seed means (shaded: $\pm1$ sd). Dotted: the TA start.}
\label{fig:ablation}
\end{figure}

\paragraph{Inside versus beyond the task-vector span.}
With the weights frozen, every module stays at $W^{(0)}_m+\sum_k(\alpha_{k,m}-\alpha^{(0)}_{k,m})\tau_{k,m}$: the student can only move within the span of the task vectors, and \method{} reduces to learning the task-arithmetic coefficients by on-policy distillation. Training the weights as well lets the student leave the span. \Cref{tab:factorial} and \cref{fig:ablation} compare the two, together with the untrained TA start and MOPD (TA init), which trains the student weights with the mixing coefficients frozen. Learning the mixing coefficients inside the span improves on the TA start by $+10.0$ and $+16.9$ average-recovery points (4B / 1.7B), but stays far below \method{} ($+51.9$ and $+62.3$), which also has the best worst-domain recovery at both scales. The step beyond the span is small but decisive: it carries only $14.6\%$ / $10.0\%$ of the displacement energy of \method{} (\cref{fig:geometry}), yet allowing it is what separates \method{} from the coefficients-only variant. Conversely, the student weight channel without learnable coefficients, MOPD (TA init), gains $+40.2$ and $+48.5$ points but leaves MedQA behind (\cref{sec:mechanism}). The coefficients thus move the student along teacher-aligned directions, the weights adapt it beyond their span, and only the combination gives the best average and worst-domain recovery. \method{} is therefore more than a learned merge: the experts supply the directions, and the distilled model is free to leave the space that any merge of them can reach. The sensitivity to the coefficient learning rate is reported in \cref{app:alphalr}.

\begin{table}[t]
\centering
\caption{Inside versus beyond the task-vector span, final checkpoint. Rows from top to bottom: the untrained TA start, mixing coefficients only (student weights frozen, confined to the span), student weights only (mixing coefficients frozen; MOPD (TA init)), and both channels (\method{}); the last two rows are three-seed means. Avg.: average recovery; Worst: worst-domain recovery.}
\label{tab:factorial}
\small
\setlength{\tabcolsep}{4pt}
\begin{tabular}{cc|cc|cc}
\toprule
& & \multicolumn{2}{c|}{Qwen3-4B} & \multicolumn{2}{c}{Qwen3-1.7B} \\
Train $W$ & Train $\alpha$ & Avg. & Worst & Avg. & Worst \\
\midrule
 & & 42.98 & 23.67 & 32.41 & 17.15 \\
\midrule
 & \checkmark & 52.98 & 35.62 & 49.31 & 27.17 \\
\checkmark & & 83.13 & 47.22 & 80.93 & 32.03 \\
\rowcolor{gray!15}\checkmark & \checkmark & \textbf{94.85} & \textbf{89.42} & \textbf{94.74} & \textbf{88.01} \\
\bottomrule
\end{tabular}
\end{table}

\paragraph{Coefficient granularity.}
\Cref{tab:granularity} compares module, layer and model granularity at the end of training. At 1.7B all three end between $94$ and $97\%$ average recovery, and layer tying ends with the lowest across-domain std ($1.5$). At 4B, model tying ($96.5$) falls within the seed range of module granularity ($93.5$--$97.2$), so the two cannot be separated, whereas layer tying ends at $73.4$, about $20$ points below both. The 1.7B layer run does not show this drop. Model tying has the highest mean at both scales, but its across-domain std depends on the scale ($3.6$ at 4B, $9.3$ at 1.7B). Module-wise granularity is the setting that combines high average recovery with a low across-domain std at both scales (average recovery $94.9$ / $94.7$, std $4.6$ / $4.8$). We therefore use it as the main setting.

\begin{table}[t]
\centering
\caption{Coefficient granularity at the final checkpoint. Avg.\ Rec.: average recovery; std: across-domain std of recovery; \#$\alpha$: number of free coefficients. Module (shaded, the main setting) is a three-seed mean.}
\label{tab:granularity}
\small
\setlength{\tabcolsep}{3.8pt}
\begin{tabular}{l>{\columncolor{gray!15}}ccc>{\columncolor{gray!15}}ccc}
\toprule
& \multicolumn{3}{c}{Qwen3-4B} & \multicolumn{3}{c}{Qwen3-1.7B} \\
\cmidrule(lr){2-4}\cmidrule(lr){5-7}
& module & layer & model & module & layer & model \\
\midrule
\#$\alpha$ & 1008 & 144 & 4 & 784 & 112 & 4 \\
Avg.\ Rec.\,$\uparrow$ & 94.9 & 73.4 & 96.5 & 94.7 & 94.4 & 97.2 \\
std\,$\downarrow$ & 4.6 & 16.6 & 3.6 & 4.8 & 1.5 & 9.3 \\
\bottomrule
\end{tabular}
\end{table}

\paragraph{Initialization.}
The advantage of \method{} over MOPD does not come from the TA start. At 1.7B we also start both methods from the base model ($\alpha_0=0$, so that $\eta_\alpha=0$ is MOPD), with otherwise identical settings. \method{} leads MOPD at every checkpoint and at both coefficient learning rates we tried ($10^{-2}$ and $3\times10^{-2}$), ending at $61.2$ and $66.2$ average recovery against $-15.7$ (\cref{app:alphalr}). As from the TA start, most of the gain comes from MedQA. The expert priors therefore help even when the student does not start at the merge: \method{} does not depend on a merged initialization to improve on MOPD, and it learns from the base student itself how much of each expert to draw on.

\paragraph{Learned coefficients.}
The learned mixture is stable: after training, the mean coefficient of each teacher agrees across the three seeds to within $0.01$, so the mixture is a structured outcome learned from the experts and the student. It also moves away from the uniform $0.25$ in a consistent direction. The medical coefficient rises the most (to $0.63$ at 4B and $0.65$ at 1.7B, against $0.33$--$0.46$ for the other three teachers; \cref{tab:alpha}), toward the domain that MOPD (TA init) fails to recover (\cref{sec:mechanism}). When the weights are frozen, the coefficients have to absorb the whole update: they move much further (std over all module coefficients $0.38$--$0.40$ versus $0.12$--$0.14$), and some module coefficients become negative. Learned jointly with the weights, they stay a moderate mixture of the experts, because the student weight channel adapts what the task-vector span cannot express. The learned coefficients are thus a reproducible and interpretable by-product of training: without any validation data, they identify the teacher that the student most needs.

\section{Conclusion}
\label{sec:conclusion}

We proposed \method{}, which treats a linear combination of the student and frozen experts as a single model and trains its weights and mixing coefficients jointly under the MOPD objective. Across four domains at 1.7B and 4B, it surpasses the final average recovery of MOPD in half the training steps and raises worst-domain recovery from below half to nearly $90\%$. The gain comes from reachability. Student weight-only updates barely move along the task vector of the lagging domain, whereas one learnable coefficient per module makes this direction reachable, and the student weight channel adapts the model beyond the task-vector span. The advantage persists without a merged initialization, and the learned coefficients consistently identify the teacher that the student most needs. More broadly, expert parameters need not be consumed once, as an initialization or a merge. Kept as learnable priors throughout distillation, they guide where the student moves, which makes expert weights, and not only expert predictions, a source of supervision for multi-teacher consolidation.

\bibliography{example_paper}

\begin{thebibliography}{35}
\providecommand{\natexlab}[1]{#1}
\providecommand{\url}[1]{\texttt{#1}}
\expandafter\ifx\csname urlstyle\endcsname\relax
  \providecommand{\doi}[1]{doi: #1}\else
  \providecommand{\doi}{doi: \begingroup \urlstyle{rm}\Url}\fi

\bibitem[Amari(1998)]{amari1998natural}
Amari, S.-i.
\newblock Natural gradient works efficiently in learning.
\newblock \emph{Neural Computation}, 10\penalty0 (2):\penalty0 251--276, 1998.

\bibitem[Arora et~al.(2018)Arora, Cohen, and Hazan]{arora2018optimization}
Arora, S., Cohen, N., and Hazan, E.
\newblock On the optimization of deep networks: Implicit acceleration by overparameterization.
\newblock In \emph{International Conference on Machine Learning (ICML)}, 2018.

\bibitem[Barres et~al.(2025)Barres, Dong, Ray, Si, and Narasimhan]{barres2025tau2}
Barres, V., Dong, H., Ray, S., Si, X., and Narasimhan, K.
\newblock $\tau^2$-bench: Evaluating conversational agents in a dual-control environment.
\newblock \emph{arXiv preprint arXiv:2506.07982}, 2025.

\bibitem[Choshen et~al.(2022)Choshen, Venezian, Slonim, and Katz]{choshen2022fusing}
Choshen, L., Venezian, E., Slonim, N., and Katz, Y.
\newblock Fusing finetuned models for better pretraining.
\newblock \emph{arXiv preprint arXiv:2204.03044}, 2022.

\bibitem[{DeepSeek-AI}(2026)]{deepseekai2026deepseekv4}
{DeepSeek-AI}.
\newblock {DeepSeek-V4}: Towards highly efficient million-token context intelligence.
\newblock \emph{arXiv preprint arXiv:2606.19348}, 2026.

\bibitem[Fu et~al.(2026)Fu, He, Zuo, Huang, Zhang, Xiao, Qian, Luo, Gao, Wang, et~al.]{fu2026rethinking}
Fu, Z., He, B., Zuo, Y., Huang, H., Zhang, J., Xiao, R., Qian, C., Luo, Q., Gao, H.-a., Wang, Y., et~al.
\newblock Rethinking on-policy distillation of large language models ii: One training example.
\newblock \emph{arXiv preprint arXiv:2609.04172}, 2026.

\bibitem[Gao et~al.(2026)Gao, Chi, Yan, Feng, Wu, Jiang, He, Ma, Zhang, and Zhou]{gao2026openmopd}
Gao, H.-a., Chi, H., Yan, Y., Feng, S., Wu, H., Jiang, Z., He, B., Ma, W.-Y., Zhang, Y.-Q., and Zhou, H.
\newblock Open-mopd: Diagnosing and fixing capability imbalance in multi-teacher on-policy distillation.
\newblock \emph{arXiv preprint arXiv:2608.19098}, 2026.

\bibitem[Gauthier-Caron et~al.(2024)Gauthier-Caron, Siriwardhana, Stein, Ehghaghi, Goddard, McQuade, Solawetz, and Labonne]{gauthiercaron2024dam}
Gauthier-Caron, T., Siriwardhana, S., Stein, E., Ehghaghi, M., Goddard, C., McQuade, M., Solawetz, J., and Labonne, M.
\newblock Merging in a bottle: Differentiable adaptive merging ({DAM}) and the path from averaging to automation.
\newblock \emph{arXiv preprint arXiv:2410.08371}, 2024.

\bibitem[Ilharco et~al.(2023)Ilharco, Ribeiro, Wortsman, Gururangan, Schmidt, Hajishirzi, and Farhadi]{ilharco2023task}
Ilharco, G., Ribeiro, M.~T., Wortsman, M., Gururangan, S., Schmidt, L., Hajishirzi, H., and Farhadi, A.
\newblock Editing models with task arithmetic.
\newblock In \emph{International Conference on Learning Representations (ICLR)}, 2023.

\bibitem[Kang et~al.(2026)Kang, Galim, Oh, Kang, Park, Kim, Lee, Kim, Tiwari, Zeng, et~al.]{kang2026asyncopd}
Kang, W., Galim, K., Oh, S., Kang, M., Park, S., Kim, D., Lee, M., Kim, M., Tiwari, R., Zeng, Y., et~al.
\newblock Asyncopd: How stale can on-policy distillation be?
\newblock \emph{arXiv preprint arXiv:2606.24143}, 2026.

\bibitem[Kim et~al.(2026)Kim, Jang, Lee, Kim, and Lee]{kim2026no}
Kim, S., Jang, C., Lee, N., Kim, B., and Lee, J.
\newblock No pain, more gain: Iterative merging for effective multi-teacher on-policy distillation.
\newblock \emph{arXiv preprint arXiv:2609.34745}, 2026.

\bibitem[{Kimi Team}(2026)]{kimiteam2026kimik3}
{Kimi Team}.
\newblock {Kimi K3}: Open frontier intelligence.
\newblock \emph{arXiv preprint arXiv:2607.24653}, 2026.

\bibitem[Li et~al.(2026)Li, Zuo, He, Zhang, Xiao, Qian, Yu, Gao, Yang, Liu, et~al.]{li2026rethinking}
Li, Y., Zuo, Y., He, B., Zhang, J., Xiao, C., Qian, C., Yu, T., Gao, H.-a., Yang, W., Liu, Z., et~al.
\newblock Rethinking on-policy distillation of large language models: Phenomenology, mechanism, and recipe.
\newblock \emph{arXiv preprint arXiv:2604.13016}, 2026.

\bibitem[Lu \& {Thinking Machines Lab}(2025)Lu and {Thinking Machines Lab}]{lu2025onpolicy}
Lu, K. and {Thinking Machines Lab}.
\newblock On-policy distillation.
\newblock \emph{Thinking Machines Lab: Connectionism}, 2025.
\newblock \doi{10.64434/tml.20251026}.
\newblock \url{https://thinkingmachines.ai/blog/on-policy-distillation}.

\bibitem[Ma et~al.(2026{\natexlab{a}})Ma, Wei, Zhao, Zhang, Xiao, Li, Yang, Gao, Wang, Li, Dong, Sui, and Luo]{ma2026mopd}
Ma, W., Wei, J., Zhao, L., Zhang, H., Xiao, B., Li, L., Yang, Q., Gao, B., Wang, Y., Li, R., Dong, J., Sui, Z., and Luo, F.
\newblock {MOPD}: Multi-teacher on-policy distillation for capability integration in {LLM} post-training.
\newblock \emph{arXiv preprint arXiv:2606.30406}, 2026{\natexlab{a}}.

\bibitem[Ma et~al.(2026{\natexlab{b}})Ma, Zhu, Jiang, and Xiao]{ma2026promptsd}
Ma, Y., Zhu, Z., Jiang, M., and Xiao, C.
\newblock One student, many teachers: Multi-task on-policy distillation via soft-prompt privileged context.
\newblock \emph{arXiv preprint arXiv:2607.18293}, 2026{\natexlab{b}}.

\bibitem[{NVIDIA}(2026)]{nvidia2026nemotron3ultra}
{NVIDIA}.
\newblock {Nemotron~3 Ultra}: Open, efficient mixture-of-experts hybrid mamba-transformer model for agentic reasoning.
\newblock \emph{arXiv preprint arXiv:2606.15007}, 2026.

\bibitem[Salimans \& Kingma(2016)Salimans and Kingma]{salimans2016weight}
Salimans, T. and Kingma, D.~P.
\newblock Weight normalization: A simple reparameterization to accelerate training of deep neural networks.
\newblock In \emph{Advances in Neural Information Processing Systems}, 2016.

\bibitem[Schulman et~al.(2017)Schulman, Wolski, Dhariwal, Radford, and Klimov]{schulman2017ppo}
Schulman, J., Wolski, F., Dhariwal, P., Radford, A., and Klimov, O.
\newblock Proximal policy optimization algorithms.
\newblock \emph{arXiv preprint arXiv:1707.06347}, 2017.

\bibitem[Shen et~al.(2025)Shen, Qin, Huang, Chen, Hu, Zhuang, Lu, Chen, and Zhao]{shen2025mot}
Shen, Z., Qin, Z., Huang, Z., Chen, H., Hu, J., Zhuang, Y., Lu, G., Chen, G., and Zhao, J.
\newblock Merge-of-thought distillation.
\newblock \emph{arXiv preprint arXiv:2509.08814}, 2025.

\bibitem[Sun et~al.(2025)Sun, Zhao, Jian, Wu, Lin, Zhu, Jia, Zhang, Wu, Ran, Hu, Jiang, Zhou, Liu, Cui, Yang, and Zhang]{sun2025tinyr1}
Sun, L., Zhao, G., Jian, X., Wu, Y., Lin, W., Zhu, Y., Jia, C., Zhang, L., Wu, J., Ran, J., Hu, S.-e., Jiang, Z., Zhou, J., Liu, W., Cui, B., Yang, T., and Zhang, X.
\newblock {TinyR1-32B-Preview}: Boosting accuracy with branch-merge distillation.
\newblock \emph{arXiv preprint arXiv:2503.04872}, 2025.

\bibitem[Wang et~al.(2026)Wang, Wang, Qin, Song, Chen, Wang, Zhou, Zhang, Zhu, Shi, et~al.]{wangevolving}
Wang, S., Wang, P., Qin, T., Song, M., Chen, Q., Wang, Q., Zhou, G., Zhang, Z., Zhu, H., Shi, D., et~al.
\newblock Evolving rollouts: Harnessing historical experience for web agent evolution in reinforcement learning.
\newblock In \emph{Forty-third International Conference on Machine Learning}, 2026.

\bibitem[{Xiaomi LLM-Core Team}(2026{\natexlab{a}})]{xiaomi2026mimov26}
{Xiaomi LLM-Core Team}.
\newblock {MiMo-V2.6}: Scaling reinforcement learning towards self-improvement.
\newblock \emph{arXiv preprint arXiv:2610.11959}, 2026{\natexlab{a}}.

\bibitem[{Xiaomi LLM-Core Team}(2026{\natexlab{b}})]{xiaomi2026mimov2flash}
{Xiaomi LLM-Core Team}.
\newblock {MiMo-V2-Flash} technical report.
\newblock \emph{arXiv preprint arXiv:2601.02780}, 2026{\natexlab{b}}.

\bibitem[Xing et~al.(2026)Xing, Wang, Gao, Li, and Tang]{xing2026trust}
Xing, X., Wang, H., Gao, B., Li, Z., and Tang, Y.
\newblock Trust region on-policy distillation.
\newblock \emph{arXiv preprint arXiv:2606.01249}, 2026.

\bibitem[Xu et~al.(2025)Xu, Li, and Zhang]{xu2025prodistill}
Xu, J., Li, J., and Zhang, J.
\newblock Scalable model merging with progressive layer-wise distillation.
\newblock In \emph{International Conference on Machine Learning (ICML)}, 2025.

\bibitem[Xu et~al.(2026)Xu, Zheng, Zhang, Yu, Ma, Zhu, Wu, Ye, Liu, Dong, Zhu, Huang, and Yu]{xu2026mopdrouter}
Xu, T., Zheng, Y., Zhang, Z., Yu, Y., Ma, C., Zhu, J., Wu, L., Ye, L., Liu, P., Dong, B., Zhu, H., Huang, R., and Yu, G.
\newblock {MOPD-Router}: Rethinking teacher routing in multi-teacher on-policy distillation.
\newblock \emph{arXiv preprint arXiv:2609.30837}, 2026.

\bibitem[Yadav et~al.(2023)Yadav, Tam, Choshen, Raffel, and Bansal]{yadav2023ties}
Yadav, P., Tam, D., Choshen, L., Raffel, C., and Bansal, M.
\newblock {TIES}-merging: Resolving interference when merging models.
\newblock In \emph{Advances in Neural Information Processing Systems (NeurIPS)}, 2023.

\bibitem[Yang et~al.(2025)]{yang2025qwen3}
Yang, A. et~al.
\newblock {Qwen3} technical report.
\newblock \emph{arXiv preprint arXiv:2505.09388}, 2025.

\bibitem[Yang et~al.(2024)Yang, Wang, Shen, Liu, Guo, Wang, and Tao]{yang2024adamerging}
Yang, E., Wang, Z., Shen, L., Liu, S., Guo, G., Wang, X., and Tao, D.
\newblock {AdaMerging}: Adaptive model merging for multi-task learning.
\newblock In \emph{International Conference on Learning Representations (ICLR)}, 2024.

\bibitem[Yang et~al.(2026{\natexlab{a}})Yang, Liu, Xie, Yang, Yang, and Lin]{yang2026gopd}
Yang, W., Liu, W., Xie, R., Yang, K., Yang, S., and Lin, Y.
\newblock Learning beyond teacher: Generalized on-policy distillation with reward extrapolation.
\newblock \emph{arXiv preprint arXiv:2602.12125}, 2026{\natexlab{a}}.

\bibitem[Yang et~al.(2026{\natexlab{b}})]{yang2026nemotroncascade2}
Yang, Z. et~al.
\newblock {Nemotron-Cascade~2}: Post-training {LLMs} with cascade {RL} and multi-domain on-policy distillation.
\newblock \emph{arXiv preprint arXiv:2603.19220}, 2026{\natexlab{b}}.

\bibitem[Yu et~al.(2024)Yu, Yu, Yu, Huang, and Li]{yu2024dare}
Yu, L., Yu, B., Yu, H., Huang, F., and Li, Y.
\newblock Language models are {Super Mario}: Absorbing abilities from homologous models as a free lunch.
\newblock In \emph{International Conference on Machine Learning (ICML)}, 2024.

\bibitem[Zhang et~al.(2024)Zhang, Albert, Rodriguez-Opazo, van~den Hengel, and Abbasnejad]{zhang2024atlas}
Zhang, F.~Z., Albert, P., Rodriguez-Opazo, C., van~den Hengel, A., and Abbasnejad, E.
\newblock Knowledge composition using task vectors with learned anisotropic scaling.
\newblock In \emph{Advances in Neural Information Processing Systems (NeurIPS)}, 2024.

\bibitem[Zhao et~al.(2026)Zhao, Xie, Liu, Huang, Pang, Chen, and Grover]{zhao2601self}
Zhao, S., Xie, Z., Liu, M., Huang, J., Pang, G., Chen, F., and Grover, A.
\newblock Self-distilled reasoner: On-policy self-distillation for large language models.
\newblock \emph{arXiv preprint arXiv:2601.18734}, 2026.

\end{thebibliography}
\bibliographystyle{icml2026}

\newpage
\appendix
\onecolumn

\section{The Distillation Objective as Implemented}
\label{app:objective}

This appendix states the exact objective behind \cref{eq:reward,eq:opd}, as used in all reported runs.

\paragraph{Rollouts and log-probabilities.}
At step $t$ the rollout engine holds the weights $\theta_t$ synchronized after step $t-1$. It samples one response per prompt at temperature $1.0$ with top-$p=1.0$ and no top-$k$. After generation, the student log-probabilities used in the reward and in the importance ratio are \emph{recomputed} by the learner with the same weights $\theta_t$ (not read from the rollout engine), and the teacher log-probabilities are computed by the routed teacher $\pi_{\theta_{d(x)}}$ on the same prefixes at temperature $1.0$.

\paragraph{Support.}
At every response position $i$, the support $\supp_i$ is formed from the \emph{student} distribution only (``student top-$k$'' strategy): the $S=16$ most probable student tokens are selected; their probabilities are renormalized among themselves; and, in descending order, a token is retained while the cumulative renormalized mass \emph{preceding} it is below $0.99$ (so the token that crosses the threshold is kept and the most probable token is always kept). The teacher distribution is evaluated on the same support; tokens outside $\supp_i$ contribute nothing. The support is rebuilt from the recomputed student log-probabilities at learner time.

\paragraph{Reward.}
With $w_i(v)=p(v)/\sum_{u\in\supp_i}p(u)$ the per-candidate reward is $r_i(v)=-w_i(v)[\log p(v)-\log q(v)]$ as in \cref{eq:reward}, with \emph{raw} (not renormalized) log-probabilities inside the bracket. Writing $Z_p=\sum_{u\in\supp_i}p(u)$ and $Z_q=\sum_{u\in\supp_i}q(u)$ for the support masses and $\tilde p,\tilde q$ for the two distributions renormalized on $\supp_i$, the summed reward is
\begin{equation}
\sum_{v\in\supp_i} r_i(v)
=-\Big[\mathrm{KL}\big(\tilde p\,\|\,\tilde q\big)+\log\tfrac{Z_p}{Z_q}\Big],
\label{eq:reward_kl}
\end{equation}
i.e.\ a support-restricted reverse KL plus a term that penalizes the student for placing more mass on its own top candidates than the teacher does. Responses that hit the generation length cap are masked out. No reward-scale normalization and no target-share or budget correction is applied in the reported runs, except for the budget balancing of Open-MOPD. The reward is used directly as the per-candidate advantage ($A_i(v)=r_i(v)$, no baseline, no group normalization), and it is recomputed at learner time from the recomputed student log-probabilities.

\paragraph{Surrogate.}
Let $\rho_i(v)=\pi_\theta(v\mid x,y_{<i})/\pi_{\theta_t}(v\mid x,y_{<i})$. The per-candidate loss is the PPO-style clipped objective with asymmetric clipping range $(1-0.2,\,1+0.28)$ and dual clipping at $3$ for negative advantages~\cite{schulman2017ppo}:
\begin{equation}
\ell_i(v)=\max\!\big(-A_i(v)\rho_i(v),\,-A_i(v)\,\mathrm{clip}(\rho_i(v),0.8,1.28)\big),
\quad
\ell_i(v)\leftarrow\min\!\big(\ell_i(v),\,-3A_i(v)\big)\ \text{if }A_i(v)<0 .
\end{equation}
Candidate losses are summed over $v\in\supp_i$ to a per-token loss, and per-token losses are averaged over all unmasked response tokens of the mini-batch (``token-mean''). Because the mini-batch equals the rollout batch and a single PPO epoch is used, each rollout batch yields exactly one optimizer step, $\theta=\theta_t$ when the gradient is formed, $\rho_i(v)=1$ up to floating-point recomputation noise, and the clipping is inactive in effect; the gradient of the surrogate then equals the gradient of \cref{eq:opd}.

\paragraph{Masking.}
Responses whose generation ended by hitting the length cap are flagged at rollout time and excluded from the loss mask entirely (all of their tokens), before token-mean aggregation. Prompts are never trained on.

\paragraph{Per-domain decomposition.}
For analysis, the learner performs the backward pass domain by domain: the token-mean loss is restricted to the tokens of domain $d$ and multiplied by that domain's share of unmasked tokens, so that the per-domain gradients sum exactly to the batch gradient $G_t$. After each domain's backward pass the per-domain coefficient gradients $\langle G^{(d)}_m,\tau_{k,m}\rangle$ are recorded, without clearing the accumulating weight gradient, and the total coefficient gradient used in \cref{eq:update} is their sum. This decomposition changes neither the weight nor the coefficient update; it only exposes the domain-wise terms discussed in \cref{sec:method:param}.

\section{Update Rule: Implementation Details and Equivalence Conditions}
\label{app:update}

\paragraph{What is stored.}
The trainable state consists of the full-precision (FP32) master copy of the student weights $(W,\phi)$, sharded across data-parallel workers, the AdamW moments for all of them, the free coefficient table $\tilde\alpha$ and its Adam moments. $B$ is never stored. The task vectors are not stored on the accelerator; the base and expert checkpoints are memory-mapped and each $\tau_{k,m}$ is formed one module at a time when needed. %

\paragraph{Order of operations within a step.}
(1)~Rollout at $\theta_t$ and teacher scoring. (2)~Learner forward/backward producing $(G,G_\phi)$, with the per-domain decomposition above. (3)~Coefficient gradients $\langle G_m,\tau_{k,m}\rangle$ are computed from the \emph{unclipped} sharded gradients, each shard contributing its partial inner product and the partial sums reduced across shards; tied coefficients sum their modules' values. (4)~The full gradient $(G,G_\phi)$ is clipped to global norm $c_W$ and AdamW (decay $0$) is stepped on $(W,\phi)$. (5)~The coefficient gradient is clipped to norm $c_\alpha$ and Adam (decay $0$) is stepped on $\tilde\alpha$. (6)~The increment $\sum_k\Delta\alpha_{k,m}\tau_{k,m}$ is added to each FP32 module shard. (7)~Both learning-rate schedulers advance. (8)~The weights are synchronized to the rollout engine. If the clipped weight-gradient norm or the coefficient-gradient norm is non-finite, steps (4) to (6) are skipped for that batch and no state changes.

\paragraph{Equivalence to optimizing $B$.}
Let $\mathcal{A}$ denote Adam with decay zero and a given learning-rate schedule. Its state after $t$ steps is a function of the gradient sequence $G_0,\dots,G_{t-1}$ only. Since $\partial\Lt/\partial B=\partial\Lt/\partial W=G_t$, running $\mathcal{A}$ on $B$ and running it on $W$ produce the same increments $\Delta^{\mathcal{A}}_t$, and $W_{t+1}=W_t+\Delta^{\mathcal{A}}_t+D\Delta\alpha_t$ is exactly $B_{t+1}+D\alpha_{t+1}$ when $W_t=B_t+D\alpha_t$. With nonzero decay the AdamW step contains $-\eta\lambda_{\mathrm{wd}}W_t$ versus $-\eta\lambda_{\mathrm{wd}}B_t$, which differ by $-\eta\lambda_{\mathrm{wd}}D\alpha_t$; the equivalence then fails, and the materialized-$W$ form would shrink the task-vector component. All reported runs use decay $0$.

\paragraph{Checkpoints.}
Every 25 steps the FP32 weights, optimizer states and the coefficient table are written; the exported model is a standard checkpoint of $(W,\phi)$ with no coefficient metadata required for inference. The coefficient trajectory $\alpha_t$ is stored separately for analysis.

\paragraph{Measured cost.}
\Cref{tab:cost_full} reports the wall-clock and the per-step time for every coefficient granularity. The additional time is in the actor update, which streams the task vectors; rollout generation is the same for all variants. Coarser tying does not reduce this cost, since every granularity reads all task vectors once per step, and the measured differences between granularities reflect how the streaming overlaps with the rest of the step.

\begin{table}[H]
\centering
\caption{Measured training cost on $8$ GPUs for all coefficient granularities. Wall-clock covers the training job from environment ready to final checkpoint; per-step numbers are 100-step means from the training logs. $\times$: ratio to MOPD (TA init).}
\label{tab:cost_full}
\small
\setlength{\tabcolsep}{3.5pt}
\begin{tabular}{llrrrr}
\toprule
& Method & Wall (h) & $\times$ & Step (s) & Update (s) \\
\midrule
\multirow{4}{*}{4B} & MOPD (TA init) & 5.26 & 1.00 & 180.6 & 14.7 \\
& \method{} module & 6.51 & 1.24 & 225.6 & 73.8 \\
& \method{} layer & 6.32 & 1.20 & 218.2 & 68.1 \\
& \method{} model & 7.50 & 1.43 & 258.9 & 90.9 \\
\midrule
\multirow{4}{*}{1.7B} & MOPD (TA init) & 4.48 & 1.00 & 153.9 & 11.1 \\
& \method{} module & 4.90 & 1.09 & 167.4 & 37.1 \\
& \method{} layer & 4.65 & 1.04 & 160.4 & 36.4 \\
& \method{} model & 5.50 & 1.23 & 188.6 & 42.0 \\
\bottomrule
\end{tabular}
\end{table}

\section{Experimental Details}
\label{app:setup}

\paragraph{Teachers.}
At 4B we use the four released IM-MOPD domain teachers. IM-MOPD does not release its 1.7B domain teachers, so we train all four from the same 1.7B reference model following its teacher training procedure: the medical and law teachers by SFT on filtered responses of Qwen3.6-35B-A3B, and the finance and instruction-following teachers by GRPO for 200 steps. On its own benchmark, each teacher is within one point of the score reported by IM-MOPD (MedQA $58.99$ versus $58.5$, CaseHOLD $70.52$ versus $69.6$, FinQA $62.66$ versus $62.4$, IFBench $46.44$ versus $46.4$).
Qwen3-4B has 36 decoder layers and Qwen3-1.7B has 28; with seven projections per layer, module granularity has $|\modset|=252$ and $196$ modules and $K|\modset|=1{,}008$ and $784$ coefficients (layer granularity: 144 and 112; model granularity: 4).

\paragraph{Data.}
Training prompts come from per-domain pools prepared with the IM-MOPD data pipeline: 13{,}761 (medical), 31{,}396 (law), 8{,}446 (finance) and 11{,}472 (IF) prompts, 65{,}075 in total.
Batches are drawn with a domain-balanced sampler (32 prompts per domain); apart from the budget balancing of Open-MOPD, no other balancing or reweighting is applied. Evaluation uses the official test splits (1{,}273, 5{,}314, 1{,}147 and 300 items for MedQA, CaseHOLD, FinQA and IFBench). A held-out validation pool is used only by IM-MOPD, for its merge decisions; none of our methods uses it for model selection.

\paragraph{Baselines.}
MOPD (SFT warm-up) starts from 62 SFT steps on teacher samples filtered and mixed with the official IM-MOPD pipeline. MOPD (TA init) differs from \method{} only in freezing the mixing coefficients $\alpha$; the coefficients-only variant additionally freezes $(W,\phi)$.

\paragraph{Training details.}
Each step uses one sampled response per prompt (temperature $1.0$, top-$p$ $1.0$), prompts up to 15{,}872 and responses up to 16{,}384 tokens; truncated responses are masked from the loss. Each training step is one rollout batch and one optimizer step, after which the weights are synchronized to the rollout engine. The full list of hyperparameters is in \cref{tab:hparams}.

\paragraph{Seeds.}
The evaluation sampling seeds (42, 43, 44) are the same for every checkpoint. The four checkpoints of one run are not independent repeats and are never pooled as such.

\paragraph{Task-vector projection.}
For \cref{fig:geometry} the displacement of every module from the TA start is projected onto the span of its four task vectors by least squares (Gram pseudo-inverse), so the projection coefficients are in the units of $\alpha$; on frozen-weight runs they match $\alpha-0.25$ to within $0.01$ in module mean. The energy share is the squared norm of the projection divided by the squared norm of the displacement, summed over modules. The per-domain coefficient gradients are the inner products $\langle\nabla_W\mathcal{L}_d,\tau_{k,m}\rangle$ recorded at every step (\cref{app:objective}).

\paragraph{Controlled interventions.}
Each run in \cref{tab:mechanism} differs from the corresponding run of \method{} or MOPD (TA init) in a single setting. The random-direction run replaces each $\tau_{k,m}$ by an independent Gaussian random tensor rescaled to the same Frobenius norm; it starts from the same TA merge and is supervised by the same teachers, so only the directions spanned by the mixing coefficients change. A further run starts MOPD from the final coefficients of \method{} composed onto the base without its weight updates (medical KL $0.031$). It keeps MedQA high (recovery $85.1$) but leaves FinQA near the TA start (recovery $32\%$) and ends at $69.8$ average recovery; the learned-$\alpha$ point on its own is a poor model (average recovery $28.8$, below the TA merge). The coefficients therefore have to be learned together with the weights. For the run with $\alpha_{\mathrm{med}}$ frozen, we report the training-side medical KL and the MedQA recovery.

\section{Coefficient Learning Rate}
\label{app:alphalr}

\begin{figure}[H]
\centering
\includegraphics[width=0.5\textwidth]{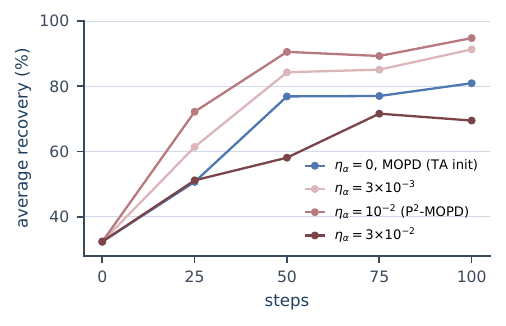}
\caption{Coefficient learning-rate sweep at 1.7B (module granularity); $\eta_\alpha{=}0$ is MOPD (TA init).}
\label{fig:alphalr}
\end{figure}

\Cref{fig:alphalr} and \cref{tab:alphalr} sweep $\eta_\alpha$ at 1.7B with everything else fixed. A three times smaller value ($3\times10^{-3}$) already gives three quarters of the gain of the main setting over MOPD (TA init) ($+10.4$ versus $+13.8$ at the end of training) and leads MOPD (TA init) at every checkpoint, so the gain is not a narrow tuning peak. $10^{-2}$ is best; a three times larger value ($3\times10^{-2}$) falls below MOPD (TA init) from step 50 on, because FinQA stays at the level of the TA start.

\begin{table}[H]
\centering
\caption{Coefficient learning-rate sweep (1.7B). Average recovery at steps 25 and 100, and its area under the learning curve (AULC; mean over the four checkpoints). Rows for $\eta_\alpha=0$ and $10^{-2}$ are three-seed means.}
\label{tab:alphalr}
\small
\begin{tabular}{lccc}
\toprule
$\eta_\alpha$ & Step 25 & Final & AULC \\
\midrule
$0$, MOPD (TA init) & 50.7 & 80.9 & 71.4 \\
$3{\times}10^{-3}$ & 61.4 & 91.3 & 80.5 \\
\rowcolor{gray!15}$10^{-2}$, \method{} & \textbf{72.2} & \textbf{94.7} & \textbf{86.7} \\
$3{\times}10^{-2}$ & 51.2 & 69.5 & 62.6 \\
\bottomrule
\end{tabular}
\end{table}

\paragraph{Start from the base model.}
For the base-model start of \cref{sec:ablation} ($\alpha_0=0$), the average recovery of \method{} with $\eta_\alpha=10^{-2}$ / $3\times10^{-2}$ against MOPD is $-7.9$ / $15.6$ versus $-20.0$ at step 25, $41.5$ / $49.0$ versus $3.0$ at step 50, $65.9$ / $64.9$ versus $26.6$ at step 75, and $61.2$ / $66.2$ versus $-15.7$ at step 100, where the Avg.\ Score is $51.6$ / $52.2$ versus $40.4$. The final MedQA recovery is $62.8$ / $98.5$ versus $-261.2$, and the final FinQA recovery is $37.7$ / $22.9$ versus $52.4$.

\section{Learned Coefficients}
\label{app:alpha}

\Cref{tab:alpha} lists the coefficients at the end of training. At both scales the medical coefficient grows the most, from $0.25$ to above $0.6$, while the other three end between $0.33$ and $0.46$.

\begin{table}[H]
\centering
\caption{Final mean coefficient per teacher of \method{} (module granularity), averaged over modules; ranges are over the three seeds where they differ. All coefficients start at $0.25$.}
\label{tab:alpha}
\small
\begin{tabular}{lcccc}
\toprule
& Med & Law & Fin & IF \\
\midrule
4B & 0.63 & 0.37 & 0.39 & 0.35--0.36 \\
1.7B & 0.65--0.66 & 0.46 & 0.36 & 0.33 \\
\bottomrule
\end{tabular}
\end{table}

\section{Hyperparameters and Matched Initialization}
\label{app:hparams}

\paragraph{Matched initialization.}
The merged start applies coefficient $0.25$ for each of the four teachers to every tensor of the model, including embeddings, normalization weights and the output head, using the IM-MOPD merge utility in FP32. \method{} and MOPD (TA init) load this identical checkpoint, and MOPD starts from $\theta_0$. The full list of training hyperparameters is in \cref{tab:hparams}.

\begin{table}[H]
\centering
\caption{Training hyperparameters of \method{} and the MOPD baselines.}
\label{tab:hparams}
\small
\begin{tabular}{lll}
\toprule
& 4B & 1.7B \\
\midrule
Optimizer steps & 100 & 100 \\
Prompts per step (per domain) & 128 (32) & 128 (32) \\
Responses per prompt & 1 & 1 \\
Rollout sampling & $T{=}1.0$, top-$p$ $1.0$ & $T{=}1.0$, top-$p$ $1.0$ \\
Max prompt / response tokens & 15{,}872 / 16{,}384 & 15{,}872 / 16{,}384 \\
Distillation support & top-$16$, nucleus $0.99$ & top-$16$, nucleus $0.99$ \\
Weight optimizer & AdamW, $\beta{=}(0.9,0.999)$, wd $0$ & same \\
$\eta_W$ (schedule) & $10^{-5}$ (5\% warm-up, constant) & $10^{-5}$ (5\% warm-up, constant) \\
Weight gradient clip $c_W$ & $1.0$ (global norm) & $1.0$ \\
Coefficient optimizer & Adam, $\beta{=}(0.9,0.999)$, wd $0$ & same \\
$\eta_\alpha$ (schedule) & $10^{-2}$ (5\% warm-up, constant) & $10^{-2}$ (5\% warm-up, constant) \\
Coefficient gradient clip $c_\alpha$ & $1.0$ & $1.0$ \\
Granularity (main) & module: $252\times4=1{,}008$ & module: $196\times4=784$ \\
Initial coefficients & $0.25$ for all four teachers & $0.25$ \\
Decoder layers / RoPE base & 36 / $10^{6}$ & 28 / $10^{6}$ \\
GPUs per run & 8 & 8 \\
Data-order seeds: \method{}, MOPD (TA init) & 0, 1, 2 & 0, 1, 2 \\
\bottomrule
\end{tabular}
\end{table}

\section{Evaluation Protocol and Endpoints}
\label{app:eval}

\paragraph{Harness.}
All scores come from the official IM-MOPD evaluation harness (frozen protocol file; same code revision for every method and endpoint). For each item three generations are sampled at temperature $1.0$ with seeds 42, 43 and 44 and the item score is averaged; the reported score is the mean over items (avg@3). IFBench is scored as prompt-level loose accuracy. Per-domain generation caps are 26{,}624 tokens (MedQA, CaseHOLD, FinQA) and 32{,}768 (IFBench); the serving engine limits the context to the model's configured 32{,}768 positions, so the effective cap for an item is the minimum of the domain cap and the remaining context.

\begin{table}[H]
\centering
\caption{Base and teacher endpoints (avg@3, \%) used for normalized recovery, measured with the same harness. The span $s_{\mathrm{teacher}}-s_{\mathrm{base}}$ is the denominator of recovery; one raw point corresponds to $100/\text{span}$ recovery points.}
\label{tab:endpoints}
\small
\begin{tabular}{llcccc}
\toprule
& & MedQA & CaseHOLD & FinQA & IFBench \\
\midrule
\multirow{3}{*}{4B} & base & 68.84 & 62.03 & 58.76 & 25.89 \\
& teacher & 79.86 & 73.57 & 75.41 & 59.22 \\
& span & 11.02 & 11.54 & 16.65 & 33.33 \\
\midrule
\multirow{3}{*}{1.7B} & base & 44.10 & 44.65 & 43.62 & 19.89 \\
& teacher & 58.99 & 70.52 & 62.66 & 46.44 \\
& span & 14.89 & 25.87 & 19.04 & 26.55 \\
\bottomrule
\end{tabular}
\end{table}

\paragraph{Recovery and aggregation.}
Per-domain recovery is $(s-s_{\mathrm{base}})/\text{span}$; it can exceed $1$ or fall below $0$. Average recovery is the unweighted mean over the four domains. Because the spans differ (\cref{tab:endpoints}), a one-point raw change moves recovery by $9.1$ points on 4B MedQA but $3.0$ points on 4B IFBench; raw means are therefore reported next to every average-recovery number.

\paragraph{Seed-paired comparisons.}
Pairing \method{} and MOPD (TA init) by seed, the step-25 gap in average recovery is $+6.17$ at 4B, with a sd of only $0.07$ across seeds against a seed sd of $\approx0.9$ for either method, and $+21.5\pm2.9$ at 1.7B. The step-50 checkpoint of \method{} exceeds the final checkpoint of MOPD (TA init) in every seed, by $+1.6$, $+4.5$ and $+21.3$ (4B) and by $+10.4$, $+11.3$ and $+7.1$ (1.7B). At 4B the mean paired gap is $+6.2$, $+7.8$, $+10.0$ and $+11.7$ at steps 25, 50, 75 and 100, with a final seed range of $+4.1$ to $+20.6$; at 1.7B it is largest at step 25 and then stays between $+12$ and $+14$.

\begin{table}[H]
\centering
\caption{IM-MOPD merge trajectory. At steps 25, 50 and 75 the recovery-gated rule merges the task vectors of the listed domains into the student; Pre and Post are the average recovery of the checkpoint immediately before and after the merge, and Step 100 is the final checkpoint. --: no merge at that step.}
\label{tab:immopd_merges}
\small
\setlength{\tabcolsep}{4pt}
\begin{tabular}{ll lcc lcc lcc c}
\toprule
& & \multicolumn{3}{c}{Step 25} & \multicolumn{3}{c}{Step 50} & \multicolumn{3}{c}{Step 75} & Step 100 \\
\cmidrule(lr){3-5}\cmidrule(lr){6-8}\cmidrule(lr){9-11}\cmidrule(lr){12-12}
Scale & Seed & Merged & Pre & Post & Merged & Pre & Post & Merged & Pre & Post & Final \\
\midrule
\multirow{3}{*}{4B} & 42 & Med, Law, IF & 58.9 & 43.2 & IF & 74.0 & 74.5 & -- & 83.9 & -- & 88.3 \\
& 43 & Med, Law & 62.3 & 58.0 & -- & 88.3 & -- & -- & 90.4 & -- & 86.6 \\
& 44 & Med, Law & 59.2 & 62.2 & -- & 83.1 & -- & -- & 85.2 & -- & 89.2 \\
\midrule
\multirow{3}{*}{1.7B} & 42 & Fin, IF & 42.3 & 46.4 & Fin & 49.2 & 47.7 & -- & 50.4 & -- & 77.4 \\
& 43 & IF & 38.7 & 44.5 & Med, Fin, IF & 49.8 & 32.7 & Fin & 56.3 & 51.7 & 67.1 \\
& 44 & IF & 38.1 & 44.9 & Med, IF & 55.7 & 47.4 & Fin & 54.1 & 57.3 & 59.0 \\
\bottomrule
\end{tabular}
\end{table}

\paragraph{IM-MOPD runs.}
IM-MOPD is trained with an effective batch of 128 and the same merged start, prompt pools, 100 steps, length caps and evaluation harness as the other methods, and with three training seeds. Its recovery-gated rule is applied at steps 25, 50 and 75 with $\delta=0.3$ and $\gamma=0.6$ (4B) or $0.4$ (1.7B), as in its released configurations; the main results use the pre-merge checkpoint at each step. \Cref{tab:immopd_merges} lists, for every seed, the domains merged at each step and the average recovery immediately before and after the merge. A merge changes the average recovery by between $-17.2$ and $+6.8$ points, and the steps that follow recover most of the larger drops. Its in-training validation (four samples per validation item every 25 steps) is part of its method and is included in its wall-clock. Because a seed also changes which domains pass the validation gate, the IM-MOPD seed variation includes the merge path.

\end{document}